\documentclass{Interspeech2024}

\interspeechcameraready

\title{InterBiasing: Boost Unseen Word Recognition through Biasing Intermediate Predictions}

\name[affiliation={1,2}]{Yu}{Nakagome}
\name[affiliation={1,2}]{Michael}{Hentschel}

\address{
  $^1$LINE WORKS Corporation, Japan\\
  $^2$NAVER Cloud Corporation, South Korea}
\email{y.nakagome@line-works.com}

\keywords{speech recognition, keywords biasing, connectionist temporal classification, self-conditioning}

\usepackage{arydshln}
\usepackage{multirow}

\begin{document}

\maketitle


\begin{abstract}
\label{sec:abst}
    
    Despite recent advances in end-to-end speech recognition methods, their output is biased to the training data’s vocabulary, resulting in inaccurate recognition of unknown terms or proper nouns. To improve the recognition accuracy for a given set of such terms, we propose an adaptation parameter-free approach based on Self-conditioned CTC. Our method improves the recognition accuracy of misrecognized target keywords by substituting their intermediate CTC predictions with corrected labels, which are then passed on to the subsequent layers. First, we create pairs of correct labels and recognition error instances for a keyword list using Text-to-Speech and a recognition model. We use these pairs to replace intermediate prediction errors by the labels. Conditioning the subsequent layers of the encoder on the labels, it is possible to acoustically evaluate the target keywords. Experiments conducted in Japanese demonstrated that our method successfully improved the F1 score for unknown words.
    
\end{abstract}


\section{Introduction}
\label{sec:intro}
In recent years, the rapid advancement of deep neural networks has led to remarkable recognition performance of end-to-end (E2E) speech recognition models, including connectionist temporal classification (CTC)~\cite{Graves06_icml}, recurrent neural network transducers~\cite{Graves12_ICMLRLW}, and attention-based encoder-decoders~\cite{chorowski2015attention,chan2016listen}.
These models are now widely used by many users in practical applications such as conference recording and AI dialogue systems.
However, the performance of these machine learning models is strongly dependent on the available training data.
As a result, recognizing rare words that are not readily available in the training data, such as internal terms, person names, and industry-specific terminology, remains a challenging problem.
In many practical scenarios, these rare terms are obtainable in advance, for example, from user names, meeting chat logs or documents, or even words registered by users.
There is a growing demand for technology that allows user customization by leveraging such information.
Furthermore, since each user has unique word requirements and new words are continually added, it is desirable that the model does not require additional training.

Traditional approaches to address these limitations have included the use of weighted finite-state transducers~\cite{zhao19d_interspeech,Huang2020ClassLA,Le2021is}.
However, these approaches require the creation and adaption of a decoding graph, which is not desirable for E2E speech recognizers that can use graph-free decoding such as models using CTC.
Moreover, alternative approaches like employing deep biasing with text embeddings~\cite{huang23d_interspeech,sudo24icassp} and integrating error correction mechanisms~\cite{Wang2022TowardsCS} have been explored.
However, it is important to note that, in text embedding methods, the representation of the model depends on its training data, which limits biasing towards the words included in the training data.

A method that neither requires model retraining nor decoding graph generation is keyword-boosted beam search (KBBS)~\cite{namkyu2022kwdboost}.
This is a practical technique that gives a bonus score if a word in the keyword list appears in the beam search hypothesis.
However, its limitation lies in its inability to award bonuses if the target keyword is not present within the search beam.
This issue is particularly pronounced in languages like Japanese, which feature multiple spellings; if the spelling of the target keywords differs from the spelling in the training data, the target keywords are less likely to appear in the search beam.
The peaky posterior distribution of CTC~\cite{Zeyer2021WhyDC} further enhances this issue.
For keyword boosting to be effective, the posterior probability of the target keyword must be a high value.

In this paper, we propose InterBiasing, a novel biasing method for Self-conditioned CTC~\cite{nozaki21_interspeech}, designed to condition intermediate layers in the acoustic model on the target keywords.
Self-conditioned CTC has achieved state-of-the-art performance in non-autoregressive E2E models~\cite{Higuchi21_asru}.
In this architecture, the CTC predictions are computed in the intermediate encoder layers, and the subsequent layers are conditioned on the predicted CTC sequence.
This process is then repeated in multiple layers.
It can be interpreted as an iterative refinement~\cite{chi2020align,higuchi2020mask,nakagome22_interspeech} of the prediction results within a single encoder.
By replacing the intermediate predictions of the target vocabulary with the correct labels and conditioning the subsequent layers on them, we can harness the framework of iterative refinement to enhance the intermediate predictions while taking the target vocabulary into consideration.
The procedure of the proposed method can be described as follows.
First, we generate speech corresponding to a list of keywords using Text-to-Speech (TTS) synthesis.
Next, this speech is input into a recognition model, which produces pairs of the correct labels and recognition errors.
Subsequently, we utilize these pairs to correct intermediate prediction errors by substituting them with the accurate labels.
By ensuring that the subsequent layers of the encoder are informed by these correct labels, we can acoustically assess the target keywords.
Our approach counter-acts CTC's peaky posterior distribution and allows the target keywords to appear in the search beam where they can be further boosted.
Moreover, since only the keywords searched from the utterance hypothesis bias the intermediate outputs, the target word can be biased regardless of the size of the keyword list.

\begin{figure*}[t]
  \centering
  \includegraphics[width=\linewidth]{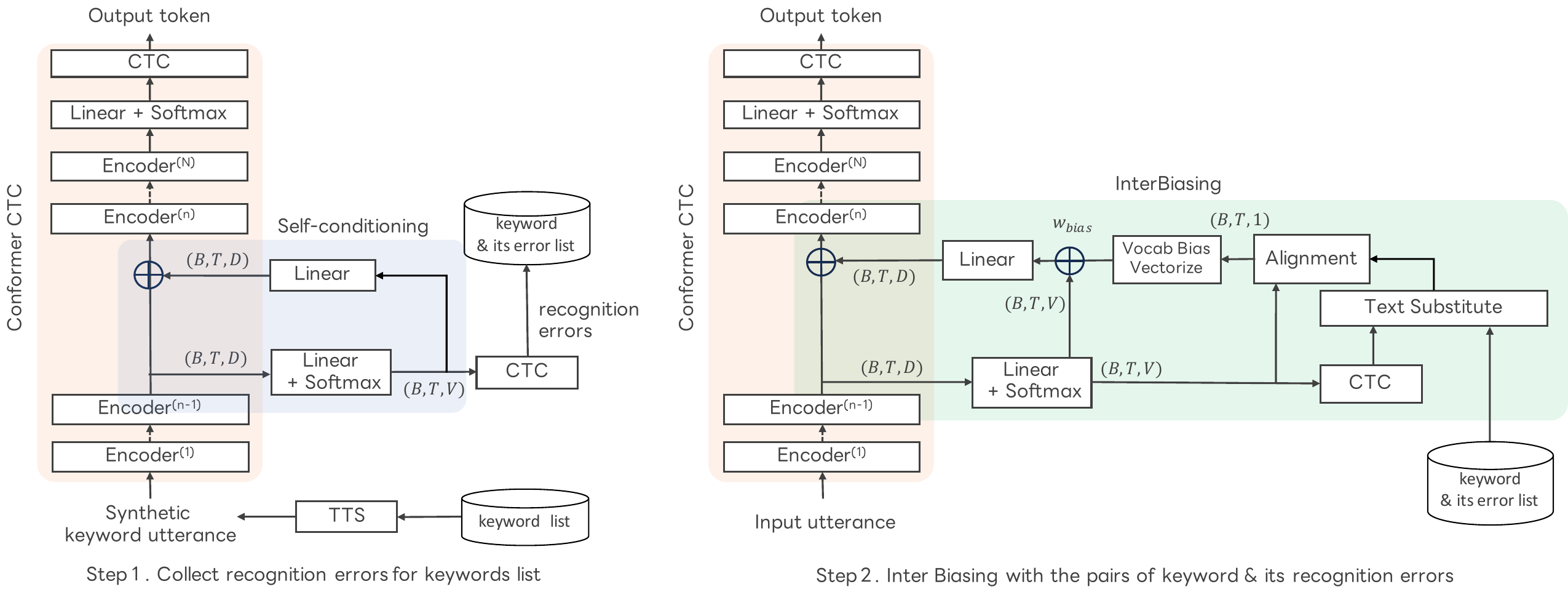}
  \caption{Overview of the proposed InterBiasing methods. The proposed method has two steps. Step 1) Speech for a keyword list is generated via Text-to-Speech (TTS), and this speech is fed into a recognition model to create pairs of correct labels and recognition error instances. Step 2) These pairs are utilized to replace intermediate prediction errors with the correct labels, and the subsequent layers infer recognition hypothesis based on these labels.}
  \label{fig:prop}
\end{figure*}

\section{Background}
\label{sec:background}

In this section, we give an overview of CTC and Self-conditioned CTC, which are the background of InterBiasing.
\subsection{Connectionist Temporal Classification}
\label{sec:ctc}
End-to-end ASR aims to model the probability distribution of a token sequence $Y=(y_l \in \mathcal{V} \mid l=1,\dots,L)$ given a sequence of $D$-dimensional audio features $X=(\mathbf{x}_t \in \mathbb{R}^D \mid t=1,\dots,T)$, where $\mathcal{V}$ is a token vocabulary.
In the CTC framework \cite{Graves06_icml}, frame-level alignment paths between $X$ and $Y$ are introduced with a special blank token $\epsilon$.
An alignment path is denoted by $\pi=\left(\pi_t \in \mathcal{V}' \mid t=1,\dots,T\right)$, where $\mathcal{V}'= \mathcal{V} \cup \{ \epsilon \}$.
The alignment path can be transformed into the corresponding token sequence by using the collapsing function $\mathcal{B}$ that removes all repeated tokens and blank tokens.
A neural network is trained to estimate the probability distribution of $\pi_t$.
We denote the output sequence of the neural network by $Z = (\mathbf{z}_t \in (0,1)^{|\mathcal{V}'|} \mid t=1,\dots,T)$, where the element $z_{t,k}$ is interpreted as $p(\pi_t = k |X)$.
The training objective of CTC is the negative log-likelihood over all possible alignment paths with the conditional independence assumption per frame, as follows:
\begin{align}\label{eq:lossctc}
    \mathcal{L}_\mathsf{ctc}(Z, Y) = - \log \sum_{\pi\in\mathcal{B}^{-1}(Y)}\prod_{t} z_{t,\pi_t}.
\end{align}
The estimated token sequence $\hat{Y}$ is obtained as follows:
\begin{align}\label{eq:ctcdec}
    \hat{Y} = \mathcal{B}(\mathsf{argmax}(Z)).
\end{align}

\subsection{Conformer Encoder}
\label{sec:conformer}
The neural network used in this paper has $N$-stacked Conformer encoders \cite{gulati20_interspeech}.
The $n$-th encoder accepts an input sequence $X^{(n-1)}$ and produces an encoded sequence of the same shape:
\begin{align}
\label{eq:encoder-output}
    X^{(n)} = \mathsf{Encoder}^{(n)}(X^{(n-1)}) \qquad (1 \le n \le N),
\end{align}
where $X^{(0)} = X$ is a subsampled sequence of input audio features.
The output sequence $Z$ is obtained by applying a linear transformation and the softmax function:
\begin{align}
\label{eq:out}
    Z = \mathsf{Softmax}(\mathsf{Linear}_{D\rightarrow |\mathcal{V}'|}(X^{(N)})),
\end{align}
where $\mathsf{Linear}_{D\rightarrow |\mathcal{V}'|}(\cdot)$ maps a $D$-dimensional vector into  a $|\mathcal{V}'|$-dimensional vector for each element of $X^{(N)}$.

\subsection{Self-conditioned CTC}
\label{sec:sc-ctc}
For regularizing the CTC model training, Intermediate CTC \cite{lee21_icassp} introduces an additional loss for output sequences of intermediate encoders.
An intermediate output sequence for the $n$-th encoder $Z^{(n)} = (\mathbf{z}^{(n)}_t \in (0,1)^{|\mathcal{V}'|}| t=1,\dots,T)$ is computed using the same linear transformation as Eq.~\ref{eq:out}:
\begin{align}
\label{eq:softmax}
    Z^{(n)} = \mathsf{Softmax}(\mathsf{Linear}_{D\rightarrow |\mathcal{V}'|}(X^{(n)})).
\end{align}
The loss for the intermediate output sequence is the same as Eq.~\ref{eq:lossctc}, and is added to the final training objective as follows:
\begin{equation}
    \mathcal{L}_\mathsf{ic} = (1-\lambda)\mathcal{L}_\mathsf{ctc}(Z,Y) + \frac{\lambda}{|\mathcal{N}|}\sum_{n \in \mathcal{N}} \mathcal{L}_\mathsf{ctc}(Z^{(n)},Y),
\end{equation}
where $\lambda \in (0,1)$ is a mixing weight and $\mathcal{N}$ is a set of encoder indices for intermediate losses.

Self-conditioned CTC \cite{nozaki21_interspeech} utilizes the intermediate output sequence for conditioning the subsequent encoders:
\begin{align}
    C^{(n)} &= \mathsf{Linear}_{|\mathcal{V}'|\rightarrow D}(Z^{(n)}), \label{eq:linear-prime}\\
    X'^{(n)} &=
        \begin{cases}
            X^{(n)} + C^{(n)} & (n \in \mathcal{N}), \\
            X^{(n)} & (n \notin \mathcal{N}),
        \end{cases}
\label{eq:selfcond}
\end{align}
 where $C^{(n)} = (\mathbf{c}^{(n)}_t \mid t=1,\dots,T)$ and $\mathsf{Linear}_{|\mathcal{V}'|\rightarrow D}(\cdot)$ maps a $|\mathcal{V}'|$-dimensional vector into a $D$-dimensional vector for each element in the input sequence.
This linear layer is shared among the intermediate layers.


\section{InterBiasing: Biasing Intermediate Predictions}
\label{sec:prop}
Figure~\ref{fig:prop} illustrates the proposed InterBiasing framework.
The proposed method substitutes misrecognition of keywords with appropriate labels in the intermediate predictions. The corrected predictions are converted to frame-level biasing features $C^{(n)}_\mathsf{Bias}$ and the encoder output~$X^{(n)}$ in Eq.~\ref{eq:encoder-output} is conditioned on them as follows:
\begin{align}
    X'^{(n)} &= X^{(n)} + C^{(n)}_\mathsf{Bias}.
\end{align}
\subsection{Collecting Recognition Errors for a Keyword List with Synthetic Audio}
To collect examples of misrecognitions of unknown keywords, we utilize the intermediate prediction results from the TTS audio for these keywords. The audio $X_{\mathsf{TTS},\kappa}$ from the set of keywords $\mathcal{K}$ is generated using the following formula:
\begin{align}
    X_{\mathsf{TTS},\kappa} = \mathsf{TTS}(\kappa), \quad \forall\kappa \in \mathcal{K}.
\end{align}
Then, we obtain the intermediate softmax probability $\hat{Y}^{(n)}_{\mathsf{TTS},\kappa}$ by applying Eq. \ref{eq:ctcdec}, \ref{eq:encoder-output} and \ref{eq:softmax} to $X_{\mathsf{TTS},\kappa}$. 
In this intermediate prediction $\hat{Y}^{(n)}_{\mathsf{TTS},\kappa}$, it is observed that the predictions in the lower encoder layers deviate from the correct labels, so we select the intermediate predictions from layers after the ${M_\mathsf{bias}}$ layer as trigger words $W_{\mathsf{trigger},\kappa}$ as follows:
\begin{align}
W_{\mathsf{trigger},\kappa}=\hat{Y}^{(n)}_{\mathsf{TTS},\kappa} \qquad (n \ge M_\mathsf{bias}).
\end{align}

\subsection{Biasing Intermediate Predictions}
If the intermediate prediction $\hat{Y}^{(n)}$ contains a word that exactly matches $W_{\mathsf{trigger,\kappa}}$, the word is replaced by ${\kappa}$ to obtain $\hat{Y}^{(n)}_{\mathsf{bias}}$.
$\hat{Y}^{(n)}_{\mathsf{bias}}$ is the sequence of the text domain, which requires conversion to a frame step alignment for conditioning.
Similar to the previous study~\cite{komatsu22_interspeech}, we obtain the alignment of $A^{(n)}_{\mathsf{bias}}$ by using the Viterbi algorithm as follows:
\begin{align}\label{eq:search}
    A^{(n)}_{\mathsf{bias}} = \textsf{Viterbi}(\hat{Y}^{(n)}_{\mathsf{bias}}, Z^{(n)})
\end{align}
The alignment of $A^{(n)}_{\mathsf{bias}}$ is then converted into a onehot vector ${Z}^{(n)}_{\mathsf{bias}}$ and we form a weighted sum with the original softmax probability $Z^{(n)}$ with bias weight $w_{\mathsf{bias}}$ as follows:
\begin{align}
    {Z}^{(n)}_{\mathsf{bias}} &= \mathsf{Onehot}(A^{(n)}_{\mathsf{bias}}), \\
    Z'^{(n)} &= (1-w_{\mathsf{bias}}) Z^{(n)} + w_{\mathsf{bias}} {Z}^{(n)}_{\mathsf{bias}}
\end{align}
Finally, the intermediate predictions are converted into the biasing features using a linear layer:
\begin{align}
    C^{(n)}_\mathsf{Bias} &= \mathsf{Linear}_{|\mathcal{V}'|\rightarrow D}(Z'^{(n)}).
\end{align}
Note that when no trigger word $W_{\mathsf{trigger,\kappa}}$ is included in $\hat{Y}^{(n)}$, the conventional Self-conditioned CTC is performed.
That is, $W_{\mathsf{trigger,\kappa}}$ that do not appear in $\hat{Y}^{(n)}$ are ignored, allowing for biasing towards $\kappa$ regardless of the size of $\mathcal{K}$.


\section{Experiments}
\label{sec:experiment}

\begin{table}
\centering
\caption{Summary of keyword set sizes and average number of characters for OOV and Non-OOV keywords in each testset.}
\label{tab:datasets_summary}
\begin{tabular}{@{}m{1cm}cccc@{}}
\toprule
& \multirow{2}{*}{Dataset} & $|\mathcal{K}|$ & Avg. char length \\
 & & OOV / Non-OOV & OOV / Non-OOV \\
\midrule
\multirow{3}{1cm}{In-domain} & CSJ 1 & 2 / 8 & 5.0 / 3.1 \\
 & CSJ 2 & 4 / 3 &  4.0 / 5.7\\
 & CSJ 3 & 23 / 8 & 4.3 / 3.1 \\ \hdashline
\multirow{3}{1cm}{Out-of-domain} & CV & 59 / 54 & 5.6 / 4.3 \\
 & JSUT & 212 / 103 & 3.7 / 2.8 \\
 & TED & 99 / 78 & 4.0 / 3.6 \\
\bottomrule
\end{tabular}
\end{table}

\begin{table*}[!ht]
\centering
\caption{CERs and F1 scores of Out-of-Vocabulary (OOV) and Non-OOV words in CSJ eval1, eval2, eval3, Common Voice, JSUT basic 5000 and TEDxJP-10K. Reported metrics are in the following format: CER / F1 of OOV words / F1 of Non-OOV words.}
\label{tab:methods_comparison}
\begin{tabular}{lcccccc}
\toprule
& \multicolumn{5}{c}{CER (\%) / OOV F1 (\%) / Non-OOV F1 (\%) } \\
Methods & CSJ eval1 & CSJ eval2 & CSJ eval3 & Common Voice & JSUT basic 5000 & TEDxJP-10K  \\ 
\midrule
\multicolumn{4}{l}{{\bf Greedy decoding}}\\
SelfCond  & \underline{4.6} / \underline{75.0} / 88.9  & \underline{3.7} / 0.0 / \underline{0.0}  & \underline{3.4} / 6.2 / 69.8 & \underline{19.0} / 18.6 / 81.4 & \underline{11.7} / 9.2 / 54.6 & \underline{16.1} / 12.0 / 85.3\\
TextSub & \underline{4.6} / \underline{75.0} / 90.1 & \underline{3.7} / 0.0 / \underline{0.0}  & 3.6 / 9.1 / 72.7 & 19.9 / 18.6 / 82.3 & 12.0 / 10.0 / 56.3 & 16.2 / \underline{13.1} / 19.4 \\
InterBiasing & \underline{4.6} / \underline{75.0} / 92.3  & \underline{3.7} / \underline{40.0} / \underline{0.0}  & 3.5 / \underline{27.4} / \underline{75.6} & \underline{19.0} / \underline{22.7} / \underline{82.9} & \underline{11.7} / \underline{13.5} / \underline{59.5} & \underline{16.1} / \underline{13.1} / \underline{85.9}\\\hdashline
\multicolumn{4}{l}{{\bf LM shallow fusion + Beam Search decoding}}\\
SelfCond  & 4.5 / \underline{82.4} / 90.1  & \underline{3.7} / \underline{0.0} / \underline{33.3} & \underline{3.4} / 9.2 / 72.7 & \textbf{17.3} / \underline{18.6} / 82.8 & \textbf{11.5} / 9.2 / 54.9 & \textbf{15.8} / 12.0 / 85.6 \\ 
TextSub & 4.5 / 77.8 / 15.8 & \underline{3.7} / \underline{0.0} / \underline{33.3} & 3.5 / 12.1 / 73.7 & 18.1 / 10.7 / 79.2 & 13.0 / 8.8 / 20.4 & 16.7 / 8.3 / 8.3 \\
InterBiasing  & 4.5 / \underline{82.4} / \underline{93.5} & \underline{3.7} / \underline{0.0} / \underline{33.3} & 3.5 / \underline{36.8} / \underline{80.9} & \textbf{17.3} / \underline{18.6} / \underline{84.0} & \textbf{11.5} / \underline{12.1} / \underline{59.8} & \textbf{15.8} /  \underline{14.2} / \underline{86.3} \\ \hdashline
\multicolumn{4}{l}{{\bf LM shallow fusion + Keyword-boosted Beam Search decoding}}\\
SelfCond  & \textbf{3.9} / \textbf{88.9} / 90.3 & \textbf{2.8} / 0.0 / \textbf{75.0} & 3.4 / 29.3 / 75.6  & \underline{17.7} / 50.9 / 85.1 & \textbf{11.5} / 33.9 / 67.5 & 15.9 / 27.7 / 86.8  \\
InterBiasing &  \textbf{3.9} / \textbf{88.9} / \textbf{94.7}  & \textbf{2.8} / \textbf{40.0} / \textbf{75.0} & \textbf{3.3} / \textbf{62.4} / \textbf{83.3} & \underline{17.7} / \textbf{52.3} / \textbf{85.8} & \textbf{11.5} / \textbf{41.5} / \textbf{70.2} & \textbf{15.8} / \textbf{33.0} / \textbf{87.4}   \\ 
\bottomrule
\end{tabular}
\end{table*}

To evaluate the effectiveness of the proposed InterBiasing, we conducted speech recognition experiments using the NeMo toolkit\footnote{https://github.com/NVIDIA/NeMo} \cite{kuchaiev2019nemo}.
The performance of the models was evaluated based on character error rates (CERs) and F1 score.
Following the conventional study~\cite{namkyu2022kwdboost}, we adopted the F1 score as our primary evaluation metric.

\subsection{Data}
We utilized a model that was trained on the CSJ corpus \cite{Maekawa2003}, which contains Japanese public speeches on academic topics.
The vocabulary $\mathcal{V}$ was a set of 3,260 character units.
We used 80-dimensional Mel-spectrogram features as input features.
SpecAugment \cite{Park19_interspeech} and Speed perturbation \cite{Ko15_interspeech} were also applied with the ESPNet recipe~\cite{watanabe18_interspeech}.

We conducted the evaluation on three in-domain (CSJ eval1, eval2, eval3~\cite{Maekawa2003}) and three out-of-domain (JSUT-basic 5000~\cite{Sonobe2017JSUTCF}, Common Voice v8.0~\cite{commonvoice2020}, TEDxJP-10K~\cite{ando2021ted10k}) test sets.
The out-of-domain test sets are representative of the common scenario in applications where the unseen user data does not match the training data in acoustic or lexical conditions.

Here, we describe the process of generating bias keywords for our evaluation experiments.
Initially, we performed speech recognition on each evaluation set using a model trained on the CSJ dataset.
By comparing the labels and recognition hypotheses, we identified words that were incorrectly recognized.
In this process, word segmentation was conducted using morphological analysis with MeCab \cite{kudo-etal-2004-applying}.
From the extracted misrecognized words, we only retained proper nouns and personal names with two characters or more using morphological labels.
In the final stage, we manually removed clear morphological analysis errors from the extracted set of bias keywords.
We classified all bias keywords into out-of-vocabulary (OOV) keywords and Non-OOV keywords based on whether they belong to the vocabulary in the CSJ training data.
The number of OOV and Non-OOV keywords in each evaluation set is shown in Table \ref{tab:datasets_summary}.

\subsection{Model Configurations}
\noindent\textbf{SelfCond:}
We used the Self-conditioned CTC model as described in Section~\ref{sec:sc-ctc}.
The number of layers $N$ was 18, and the encoder dimension $D$ was 512.
The convolution kernel size and the number of attention heads were 31 and 8, respectively.
The model was trained for 50 epochs, and the final model was obtained by averaging model parameters over 10 best checkpoints in terms of validation cer values.
The effective batch-size was set to 120.
The Adam optimizer~\cite{Kingma14_iclr} with $\beta_1 = 0.9$, $\beta_2=0.98$,
the Noam Annealing learning rate scheduling~\cite{Vaswani17_NIPS} with 1k warmup steps were used for training.
Self-conditioning are applied at every layer ($\mathcal{N} = \{1,2,...,17\}$).

\noindent\textbf{TextSub (Text Substitution)}:
A simple text substitution process was applied to the final recognition hypotheses using the pairs of keywords and trigger lists.

\noindent\textbf{InterBias:}
To generate trigger words, we utilized synthetic speech via the in-house TTS system.
We then processed this synthesized speech through the above mentioned SelfCond model, using the results from greedy decoding in the intermediate layers (${M_\mathsf{bias}}$ = $3$).
The bias weight $w_{\mathsf{bias}}$ was set to 0.9. 
InterBiasing is applied at layer indices ($\mathcal{N} = \{1,2,...,17\}$).

\noindent\textbf{Beam Search decoding:}
In the LM shallow fusion, a 10-gram Ngram was trained using the text corpus from the speech training data with KenLM~\cite{heafield-2011-kenlm}.
Beam size, LM weight, and length penalty were set to 10, 0.5, and 0.2, respectively, optimized using the CSJ dev set.
The weight of KBBS was set to $3.0$.

\subsection{Results}
\label{sec:results} 
Table~\ref{tab:methods_comparison} summarizes the experimental results.
First, in greedy decoding, it can be seen that the TextSub significantly degraded the F1 score of Non-OOV words in TEDxJP-10k.
The TextSub frequently overcorrected recognition results and was less likely to hit the trigger words since they were not always present in the final outputs.
In contrast, the InterBiasing had more chances to hit trigger words because text substitution was applied for intermediate predictions on multiple layers.
Compared to SelfCond and the TextSub, the experimental results show a large improvement of F1 scores of the OOV words, and less degradation of CERs with the proposed InterBiasing.
It should be noted that the large fluctuations in the results for CSJ eval1 and eval2 scores have little significance due to the low number of keywords being evaluated, as shown in Table~\ref{tab:datasets_summary}.

Next, in beam search decoding with LM shallow fusion, we also observed that the proposed InterBiasing improved on the F1 scores of SelfCond on many evaluation sets.
As in the greedy decoding, especially for OOV keywords in CSJ eval3, InterBiasing significantly improved the OOV and Non-OOV recognition performance with beam search compared with SelfCond.
The performance of TextSub tended to degrade similarly to that observed with greedy decoding.

Finally, in LM shallow fusion and KBBS decoding, KBBS remarkably boosted the OOV and Non-OOV recognition performance of SelfCond.
However, OOV F1 scores of SelfCond remained low.
It was confirmed that the combination of the proposed InterBiasing and KBBS further boosted the keyword recognition of the SelfCond and achieved the best performance on all evaluation sets.
In particular, the recognition performance of OOV words was improved.
This appears to stem from InterBiasing increasing the acoustic score of the keywords, causing the keywords to appear in the hypothesis of KBBS.

\subsection{Analysis of the Relationship Between Beam Size and Keyword Recognition Rate}
\begin{figure}[t]
  \centering
  \includegraphics[width=\linewidth]{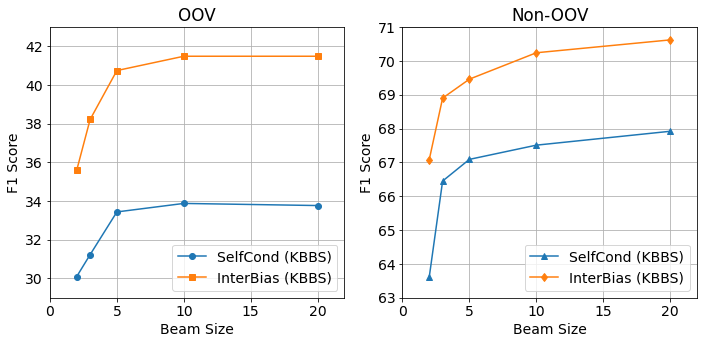}
  \caption{F1 scores of Out-of-Vocabulary (OOV) and Non-OOV words in JSUT basic 5000 with different beam sizes. LM shallow fusion and Keyword-boosted Beam Search (KBBS) were utilized. Beam size was set to 2, 3, 5, 10, 20, respectively.}
  \label{fig:beam_size}
\vspace{-10pt}
\end{figure}

In this section, we report an analysis of how the beam size impacts the keyword recognition performance.
Figure \ref{fig:beam_size} shows the F1 scores of SelfCond and InterBiasing for various beam sizes of KBBS on the JSUT basic 5000.
For the OOV keywords, SelfCond improved performance with increasing beam size from 2 to 10, but did not improve F1 scores when the beam size was further increased.
This indicates that the acoustic scores of many OOV words output from SelfCond were too low to appear in the beam search hypothesis even when the beam size was increased.
Using InterBiasing with a beam size of only 2, the keyword recognition rate was already superior to SelfCond with a beam size of 10.
Furthermore, when the beam size was increased to 10, the performance of InterBiasing was further improved.
Therefore, it can be seen that InterBiasing enhanced the acoustic score of OOV words, making these words appear more frequently within the beam search hypothesis.
As a result, the effect of KBBS is likely enhanced.
Similarly, for Non-OOV words, we observed that high keyword recognition rates were achieved even when using smaller beam sizes.

\section{Conclusions}
\label{sec:conclusions}
In this paper, we propose a method to improve the speech recognition performance of unknown words without additional training by effectively augmenting the intermediate predictions of the acoustic encoder with a keyword list and integrating it into the subsequent network layers. This approach allows for acoustic analysis of the keywords in the acoustic encoder, thereby improving the recognition performance of unknown words while minimizing side effects such as over-boosting. Experimental results in Japanese confirmed that the proposed method enhances the recognition performance of unknown words.

\bibliographystyle{IEEEtran}
\bibliography{mybib}

\end{document}